\documentclass{article}

\usepackage{arxiv}

\usepackage[utf8]{inputenc} % allow utf-8 input
\usepackage[T1]{fontenc}    % use 8-bit T1 fonts
\usepackage{hyperref}       % hyperlinks
\usepackage{url}            % simple URL typesetting
\usepackage{booktabs}       % professional-quality tables
\usepackage{amsfonts}       % blackboard math symbols
\usepackage{nicefrac}       % compact symbols for 1/2, etc.
\usepackage{microtype}      % microtypography
\usepackage{lipsum}		% Can be removed after putting your text content
\usepackage{graphicx}
\usepackage{natbib}
\usepackage{doi}

\usepackage{lineno,hyperref}
\usepackage{soul}
\usepackage{url}
\usepackage{amsmath}
\usepackage{amsthm}
\usepackage[linesnumbered, ruled, vlined]{algorithm2e}
\usepackage{multirow}
\usepackage{caption}
\usepackage{subcaption}
\modulolinenumbers[5]

\title{ACDC: Online Unsupervised Cross-Domain Adaptation\thanks{Paper under consideration for publication at Elsevier Journal}}

\author{Marcus de Carvalho\\
	School of Computer Science and Engineering\\
	Nanyang Technological University\\
	50 Nanyang Ave, Singapore 639798 \\
	\texttt{marcus.decarvalho@ntu.edu.sg} \\
	\And
    Mahardhika Pratama\\
	School of Computer Science and Engineering\\
	Nanyang Technological University\\
	50 Nanyang Ave, Singapore 639798 \\
	\texttt{mpratama@ntu.edu.sg} \\
	\And
	Jie Zhang\\
	School of Computer Science and Engineering\\
	Nanyang Technological University\\
	50 Nanyang Ave, Singapore 639798 \\
	\texttt{zhangj@ntu.edu.sg} \\
	\And
	Edward Yapp\\
	Singapore Institute of Manufacturing Technology\\
	2 Fusionopolis Way, Singapore 138634 \\
	\texttt{edward\textunderscore yapp@simtech.a-star.edu.sg} \\
}

\date{}

\renewcommand{\shorttitle}{ACDC: Online Unsupervised Cross-Domain Adaptation}

\hypersetup{
pdftitle={ACDC: Online Unsupervised Cross-Domain Adaptation},
pdfsubject={cs},
pdfauthor={Marcus de Carvalho, Mahardhika Pratama, Jie Zhang, Edward Yapp},
pdfkeywords={transfer learning, unsupervised domain-adaptation, online learning, life-long machine learning, adversarial-learning, covariate shift, asynchronous drift},
}

\begin{document}
\maketitle

\begin{abstract}
	We consider the problem of online unsupervised cross-domain adaptation, where two independent but related data streams with different feature spaces -- a fully labeled source stream and an unlabeled target stream -- are learned together. Unique characteristics and challenges such as covariate shift, asynchronous concept drifts, and contrasting data throughput arises. We propose ACDC, an adversarial unsupervised domain adaptation framework that handles multiple data streams with a complete self-evolving neural network structure that reacts to these defiances. ACDC encapsulates three modules into a single model: A denoising autoencoder that extracts features, an adversarial module that performs domain conversion, and an estimator that learns the source stream and predicts the target stream. ACDC is a flexible and expandable framework with little hyper-parameter tunability. Our experimental results under the prequential test-then-train protocol indicate an improvement in target accuracy over the baseline methods, achieving more than a 10\% increase in some cases.
\end{abstract}

% keywords can be removed

\section{Introduction}

Data mining and machine learning technologies have already achieved significant success in many data stream fields, including classification, regression, and clustering (\cite{gama2010knowledge}). However, many incremental learning algorithms work well only under a common assumption: the training and test data are drawn from the same distribution. As a result, most statistical models need to be rebuilt from scratch using newly collected training data when the distribution changes. It is expensive or impossible to re-collect the needed training data and rebuild the models in many real-world applications. Traditional transfer learning methods address complete domain adaptation problems for stationary data (\cite{TLsurvey}), but it assumes training data in the new domain is given a priori. Therefore, online unsupervised cross-domain adaptation is desirable where training samples emerge sequentially, being a challenge yet to be fully explored.

There are many cases of streaming processes where online unsupervised cross-domain adaptation can truly be beneficial. For example, the rapid development of the Internet of Things generates massive data streams and requires extra attention from the streaming mining field. An initial predictive model can be built upon a single application instance where the true class labels are usually fed back by an oracle or another offline algorithm. However, repeating this building process for each new instance is cumbersome and costly, making automatic distribution alignment between a previously labeled source stream and a new unlabeled target stream more favorable. This practical case leads to five challenges whenever a new stream model is introduced: 1) {\it Scarcity of labeled samples} on the new unlabeled stream (\cite{scarcitylabeleddata}); 2) {\it Different feature space} or {\it different marginal probability distribution} (\cite{SUN201584}) from the initial predictive model; 3) {\it Covariate shift} (\cite{WU2021108007}), as data are drawn from a different distribution; 4) {\it Asynchronous drift} (\cite{charconceptdrift}), as both streams suffer from independent drifts through time; 5) {\it Contrasting throughput}, as streams generate samples on a different speed.

Furthermore, an online model must perform under the {\bf prequential test-then-train} approach (\cite{Gama2013}). Existing approaches in online transfer learning (\cite{MSC,FUSION,ATL, OTL1}) can effectively work with scarce labels in the target domain, detect drifts, adapt to changes in the data distribution over time, and handle contrasting throughput. However, these approaches assume that the source and target streams share the same domain, i.e., they perform supervised or semi-supervised domain adaptation as some information regarding the target domain is initially known. On the other hand, online unsupervised cross-domain adaptation aligns the distribution of multiple related data streams without previous knowledge requirements. To the best of our knowledge, (\cite{onlineHeterogeneousTexas,COMC}) are the only works until this moment that solve these five challenges. However, their solutions are based on support vector machines (SVM), which were already demonstrated to have limitations in dealing with high-dimension problems compared to solutions based on neural network (\cite{domingos2020model}).

This paper presents an adversarial unsupervised cross-domain adaptation framework with a complete self-evolving neural network structure that handles multiple data streams, a setting composed of two types of non-stationary data streams in different domains with different feature spaces. A stream with plentiful labeled data is referred to as the {\it source stream}, while another independent process referred to as the {\it target stream} generates unlabeled data.

We propose the novel autonomous cross-domain conversion (ACDC) framework to address these challenges by utilizing a three-module framework that encapsulates a single and straightforward model: 1) A {\bf denoising autoencoder} (DAE) (\cite{DAE}) as a generative feature extractor; 2) {\bf domain-adversarial adaptation} (DAA) network (\cite{DANN}), which forces the DAE to align the latent feature distributions of the two domains; 3) And a {\bf discriminator} (DISC), which fits the source stream to predict the target stream. All three modules accommodate an independent {\bf self-evolving structure}, enabling them to grow and prune nodes autonomously. This dynamic structure empowers every module to actively react to changes in both source and target distributions, even in the face of the asynchronous drift rate, while successfully performing online unsupervised cross-domain adaptation solving the previously mentioned five challenges.

To be more specific, ACDC deals with each challenge as following:
\begin{itemize}
    \item{\it Scarcity of labeled samples}: Handled by the domain latent invariant space in DAE and the final prediction by DISC;
    \item{\it Different feature space}  or {\it different marginal probability distribution}: Tackled by DAA, which forces an aligned feature distribution into DAE latent space;
    \item{\it Covariate shift}: Resolved mainly by the domain adaptation procedure in DAA;
    \item{\it Asynchronous drift}: Addressed by all module's dynamic structures;
    \item{\it Contrasting throughput}: Unfold by the ACDC algorithm, which pairs and permute incoming samples into processing sliding windows.
\end{itemize}
We evaluate our framework on real-world datasets and compare the results with baseline methods, which indicate more than 10\% improvement in target accuracy in some cases.

{\bf Contributions:}
\begin{itemize}
    \item A novel framework named ACDC which highlights neural network as a solution for the online unsupervised cross-domain adaptation problem;
    \item A fully autonomous data-driven structure that can grow and prune nodes on the three training phases;
    \item The usage of a domain-adversarial bias-variance trade-off to adapt the discriminator to possible concept drifts;
    \item The integration of a domain-adversarial network learning an online unsupervised cross-domain configuration;
    \item Source-code is made publicly available for further study\footnote{ACDC source-code: \url{https://github.com/Ivsucram/ACDC}}.
\end{itemize}

\section{Related Work}

Transfer learning is typically defined under the offline setting assumption (\cite{TLsurvey}) where a model developed for a task is reused to improve the learning of another task. Domain adaptation (DA) is a specific transfer learning problem where the source and target tasks are the same; however, their domain may differ. DA is divided into three categories accordingly to the target data availability: supervised DA (SDA) (\cite{PR-supervised-domain-adaptation}), semi-supervised DA (SSDA) (\cite{PR-semi-supervised-domain-adaptation}) and {\bf unsupervised DA} (UDA) (\cite{PR-unsupervised-domain-adaptation-1, PR-unsupervised-domain-adaptation-2, PR-unsupervised-domain-adaptation-3, PR-unsupervised-domain-adaptation-4}). According to (\cite{ben2010theory}), a good representation for cross-domain transfer is one for which an algorithm cannot learn to identify the domain of origin of the input sample.

Neural networks have also been applied to transfer learning because of their power in learning high-level features (\cite{domain-adaptation-large-sentiment}). However, the direct application of conventional neural networks for online analytics has their fixed and static structure as a limitation (\cite{gama2014}), making them unable to adapt to the dynamic and evolving characteristics of the data streams. This has led to the development of pruning, regularization, parameter prediction, and many other approaches which are more suitable for online learning.

Online unsupervised domain adaptation, or online UDA, is appropriate for real-world problems where data arrive sequentially. Data streams are often generated by non-stationary distributions, which are susceptible to concept drifts (\cite{charconceptdrift}). Furthermore, when dealing with two or more streams from different domains, covariate shift (\cite{WU2021108007}), different feature space (\cite{SUN201584}) and contrasting throughput are also expected. Online algorithms use adaptation mechanisms to deal with these challenges. For example, active drift detection trigger adaptation mechanism such as the creation of new models from scratch and re-training of the model (\cite{Gama04learningwith, CD-MOA}). Contrariwise, passive drift detectors continuously adapt the model to any drifts that the streams may manifest.

Existing online domain adaptation approaches are divided into two major categories: single-domain and cross-domain.

\subsection{Single-domain online domain adapatation}

Single-domain online domain adaptation solutions can perform SDA, SSDA, or UDA, however in a single domain, i.e., the source and target streams have the same domain, where usually the target is a subset of source or vice-versa. There are multiple single domain solutions with both active and passive drift detectors.

MSC (\cite{MSC}), MSCRDR (\cite{MSCRDR}), and FUSION (\cite{FUSION}) are ensemble-based solutions built over SVM as their main classifier. To actively detect drifts, MSC uses KMM (\cite{KMM}), MSCRDR uses its own method, and FUSION uses DMM (\cite{Gama04learningwith}).

Melanie (\cite{Melanie}) and MARLINE (\cite{MARLINE}) are built using random forests, which is also an ensemble-based solution. While Melanie uses DMM as its drift detector, MARLINE uses HDDM (\cite{HDDM}). Both SVM and random forests are unable to learn high-level features from the data when compared to neural networks (\cite{domingos2020model}).

ATL and OTL are single-domain solutions as well. While an active drift detector drives ATL, OTL uses a passive drift detector and continuously adapts the model to the possible distribution changes. ATL (\cite{ATL}) presents a self-evolving neural network structure with two modules: a feature extractor and a discriminator. ATL relies on the Kullback–Leibler divergence (KL) to align the source and target distributions within the feature extractor, while its discriminator has no knowledge of the target distribution during its processes. OTL (\cite{OTL2}) constantly updates its weight with half information from the source stream and half information from the target stream, aiming to build an intermediate domain-invariant representation of their distributions.

OTL is an SDA solution, as it requires some label information in the target stream. MSC, MSCRDR, and FUSION are UDA solutions. However, they require a warm-up period to initialize their ensemble of classifiers. ATL, Melanie, and MARLINE are UDA solutions without the need for warm-up or other pre-initialization procedure.

\subsection{Cross-domain online domain adaptation}

Contrary to single-domain online domain adaptation, cross-domain online transfer learning solutions can perform SDA, SSDA, or UDA among two or more related streams, where the source and the target have different domains but common tasks. If the different stream domains have mismatched feature spaces, pre-processing can be applied to the streams before input them into the models.

MSDA and COMC (\cite{onlineHeterogeneousTexas, COMC}) are examples of cross-domain online transfer learning approaches. Both are UDA solutions that require a warm-up initialization period, use SVM as their main classifiers, and present their own active drift detection methods. 

Finally, the proposed ACDC also fits in this category, being an online unsupervised cross-domain adaptation that puts forwards neural networks as a high-dimension domain-invariance solution.

\section{Problem Formulation}

The risk according to the distribution $\mathcal{D}_\mathcal{S}$ that a hypothesis $h : \mathcal{X} \rightarrow \mathcal{Y}$ disagrees with a labelling function $f$ is defined as in Equation \eqref{eq:hypothesis}. We use the  shorthand $\epsilon_\mathcal{S}(h) = \epsilon_\mathcal{S}(h, f_\mathcal{S})$ to refer to the source error of a hypothesis. In order to measure the generalization performance in $\mathcal{D}_T$ of a model trained in $\mathcal{D}_S$, we can bound its target error $\epsilon_\mathcal{T}(h)$ in terms of the source error, as in Equation \eqref{eq:ben-david-1}, where $d_1$ is the $L^{1}$ or variation divergence.

\begin{equation} \label{eq:hypothesis}
\epsilon_\mathcal{S}(h, f) = E_{x \sim D_\mathcal{X}} \Big[\big|h(x) - f(x)\big|\Big]
\end{equation}

\begin{equation} \label{eq:variation_divergence}
d_1(\mathcal{D},\mathcal{D'}) = 2 \underset{B \in \mathcal{B}}{sup}|Pr_{\mathcal{D}}[B] - Pr_{\mathcal{D'}}[B]|
\end{equation}

\begin{equation}
\label{eq:ben-david-1}
    \epsilon_\mathcal{T}(h) \leq \epsilon_\mathcal{S}(h) + d_1(\mathcal{D}_S,\mathcal{D}_T) + min\Big[\big|E_{D_\mathcal{S}}(f_\mathcal{S}, f_\mathcal{T}), E_{D_\mathcal{T}}(f_\mathcal{S}, f_\mathcal{T})\big|\Big]
\end{equation}

In every transfer learning problem, the target domain accuracy of a classifier is bounded by its source error and the divergence between source and target distributions.

We expand the transfer learning general challenge to formalize the problem of online unsupervised cross-domain adaptation as follows. Let $\{(x_{\mathcal{S}}^{(i)}, y_{\mathcal{S}}^{(i)})\}_{i=1}^{n_\mathcal{S}}$ be a set of labeled instances of size $n_\mathcal{S}$ from a non-stationary stream generated from the source domain $\mathcal{D}_\mathcal{S}$, where $x_{\mathcal{S}}^{(i)} \in \mathcal{X}_\mathcal{S}$ and $y_{\mathcal{S}}^{(i)} \in \mathcal{Y}_\mathcal{S}$.  Similarly, let $\{(x_{\mathcal{T}}^{(i)})\}_{i=1}^{n_\mathcal{T}}$ be a set of unlabeled instances of size $n_\mathcal{T}$ from another independent non-stationary stream generated from the target domain $\mathcal{D}_\mathcal{T}$, where $x_{\mathcal{T}}^{(i)} \in \mathcal{X}_\mathcal{T}$, depicting {\it scarcity of labeled samples}. If $\mathcal{X}_\mathcal{S} \neq \mathcal{X}_\mathcal{T}$ then the domains present {\it different feature space}, requiring a transformation such that we have a {\it different  marginal  probability  distribution}, i.e. $\mathcal{X}_\mathcal{S} = \mathcal{X}_\mathcal{T}$, but $P_\mathcal{S}(x) \neq P_\mathcal{T}(x)$ and $P_\mathcal{S_{\mathcal{I}}}(y|x) = P_\mathcal{T_{\mathcal{I}}}(y|x)$.

The goal is to construct a classifier that uses $\mathcal{X}_\mathcal{S} \in {\rm I\!R}^{n_\mathcal{S} \times u}$, $\mathcal{Y}_\mathcal{S} \in {\rm I\!R}^{n_\mathcal{S} \times m}$ and $\mathcal{X}_\mathcal{T} \in {\rm I\!R}^{n_\mathcal{T} \times u}$ to predict the class label $\hat{y}_\mathcal{T}^{(i)} \in \hat{\mathcal{Y}}_\mathcal{T} \in {\rm I\!R}^{n_\mathcal{T} \times m}$ of $x_{\mathcal{T}}^{(i)} \in \mathcal{X}_\mathcal{T}$.

Both source and target distributions are generated from independent non-stationary processes at different speeds yielding {\it contrasting throughput}, and can change overtime due to {\it covariate shift}, i.e. $P_\mathcal{S}(\mathcal{X},\mathcal{Y})_{t_{\mathcal{S}}} \neq P_\mathcal{S}(\mathcal{X},\mathcal{Y})_{{t_{\mathcal{S}}}+1}$,  $P_\mathcal{T}(\mathcal{X},\mathcal{Y})_{{t_{\mathcal{T}}}} \neq P_\mathcal{T}(\mathcal{X},\mathcal{Y})_{{t_{\mathcal{T}}}+1}$. Finally, {\it asynchronous drift} occurs when $t_{\mathcal{S}} \neq t_{\mathcal{T}}$ causing shifts in the source and target domains to happen at different timestamps.

\section{ACDC}

\begin{algorithm}[ht]
\caption{ACDC}
\label{alg:algorithm}
\KwIn{Source stream $\mathcal{S}$, target stream $\mathcal{T}$}
\KwOut{Target stream labels predicted $\hat{\mathcal{Y}}_\mathcal{T}$}
\While{exist samples in $\mathcal{S}$ and $\mathcal{T}$}{
    Read $N_m$ incoming samples from $\mathcal{S}$, $\mathcal{T}$ to $W_\mathcal{S}$, $W_\mathcal{T}$; \\
    DISC predicts $W_\mathcal{T} \rightarrow \hat{\mathcal{Y}}_\mathcal{T}$; \\
    Handle contrasting throughput by pairing and permuting $W_\mathcal{S}$, $W_\mathcal{T}$ to $W_\mathcal{S}'$, $W_\mathcal{T}'$; \\
    \For{$i\leftarrow 1$ \KwTo $\kappa$}{
    \ForEach{$x_s$, $y_s$, $x_t$ \textbf{in} ($W_\mathcal{S}'$, $W_\mathcal{T}'$)}{
        \lIf*{i == 1}{
            \Begin(Adaptation: \textbackslash \textbackslash Sub-section \ref{sec:module_adaptation}) {
                DAE assess $x_s \rightarrow x_t$, $x_t \rightarrow x_s$; \\
                DAA assess $x_s \rightarrow 0$, $x_t \rightarrow 1$; \\
                DISC assess $x_s \rightarrow y_s$;
            }
        }
        \Begin(Learning: \textbackslash \textbackslash Sub-Section \ref{sec:parameter-learning}){
            DAE fit $x_s \rightarrow x_t$, $x_t \rightarrow x_s$ \eqref{eq:loss-dae}; \\
            DAA fit $x_s \rightarrow 0$, $x_t \rightarrow 1$ \eqref{eq:loss-da}; \\
            DISC fit $x_s \rightarrow y_s$ \eqref{eq:loss-disc};
        }
        }
    }
}
\end{algorithm}

\begin{figure}[ht]
\centering
\centerline{\includegraphics{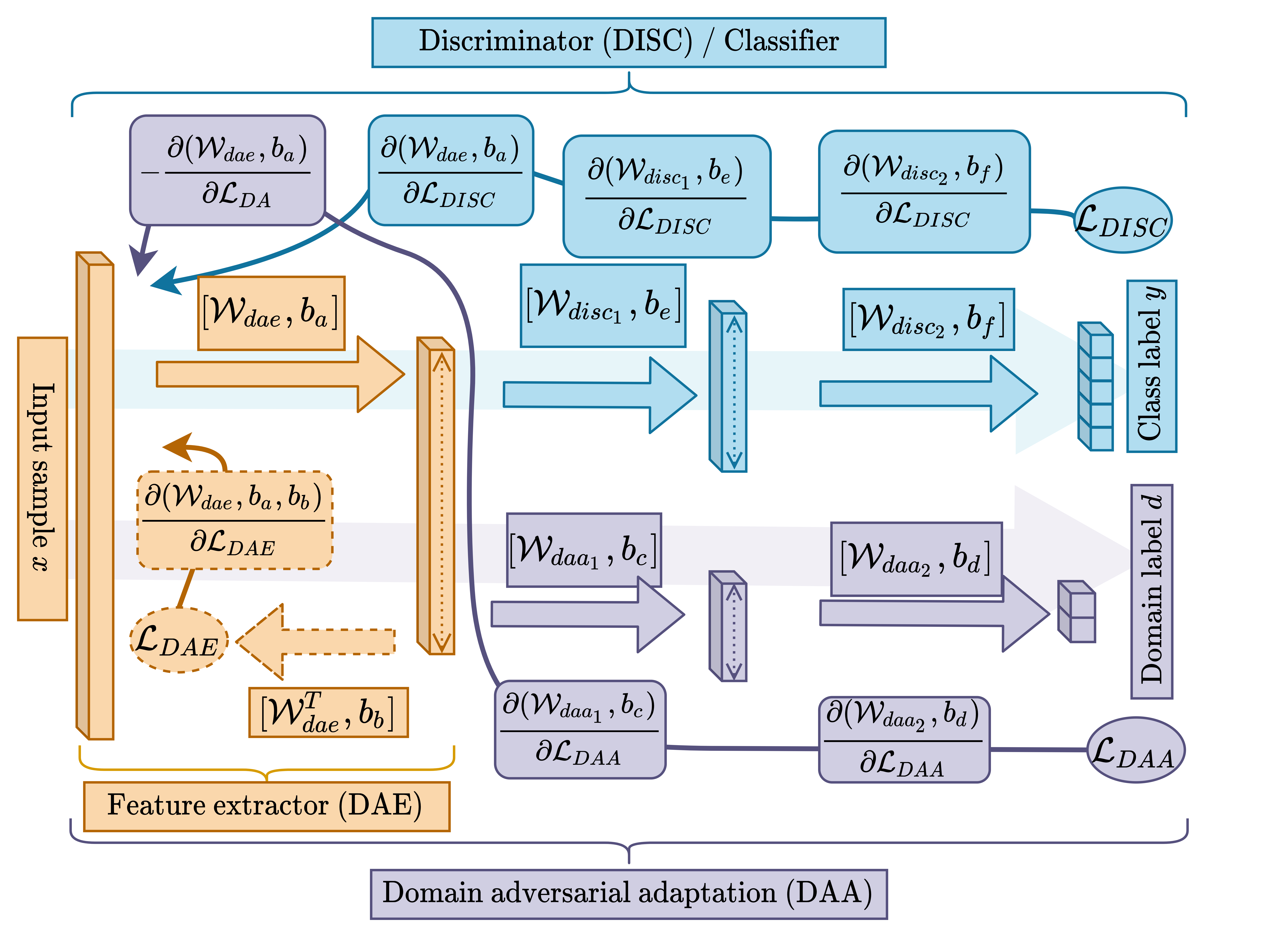}}
\caption{ACDC proposed architecture.}
\label{fig:acdc-architecture}
\end{figure}

ACDC is an online unsupervised cross-domain adaptation framework that leverages adversarial training. Adversarial UDA methods extract domain-invariant representations via deep networks and usually have a good target-domain accuracy (\cite{Sankaranarayanan_2018_CVPR}). ACDC leverages its domain-adversarial adaptation module DAA to gather information from both source and target streams, actively detecting subtle distribution changes and signalizing its discriminator DISC about possible drifts while achieving a domain-invariant feature extraction at DAE. ACDC performs all of these in a single evolving model.

An overview of the ACDC structure and learning scheme is depicted in Figure \ref{fig:acdc-architecture} while Algorithm \ref{alg:algorithm} portrays its procedural. ACDC learning procedure encapsulates three modules: {\it feature extraction}, {\it domain-adversarial adaptation} and {\it discriminator}. All three modules cover two main phases: {\it adaptation}, which grows and prunes nodes, and {\it learning}, the standard neural network feed-forwarding and back-propagation procedures.

\subsection{Parameter learning}\label{sec:parameter-learning}

The goal of ACDC is to minimize the following loss function:

\begin{equation}\label{eq:loss-acdc}
\begin{split}
 \mathcal{L}_{ACDC} = \mathcal{L}_{DAE} + \mathcal{L}_{DAA} + \mathcal{L}_{DISC}
\end{split}
\end{equation}

\noindent where:

\begin{equation}\label{eq:loss-dae}
    \mathcal{L}_{DAE} = \mathcal{L}_{mse}(\hat{x}_s,x_t) + \mathcal{L}_{mse}(\hat{x}_t,x_s)
\end{equation}

\begin{equation}\label{eq:loss-da}
    \mathcal{L}_{DAA} = (\theta_{daa} - \theta_{dae}) \times \Big(\mathcal{L}_{log}(d'_s,0) + \mathcal{L}_{log}(d'_t, 1)\Big)
\end{equation}

\begin{equation}\label{eq:loss-disc}
    \mathcal{L}_{DISC} = \mathcal{L}_{log}(\hat{y}_s, t_s)
\end{equation}

$\mathcal{L}_{mse}$ and $\mathcal{L}_{log}$ are respectively the mean-squared error and the multi-class logarithmic loss, $\hat{x}_s$ and $\hat{x}_t$ are the reconstructions of source and target samples, $d'_s$ and $d'_t$ are the binary predictions signalizing the source or target domains origin, and $\hat{y}_s$ is the model prediction to the source samples. Figure \ref{fig:acdc-architecture} presents an overview of ACDC's  weights and biases\footnote{For brevity of notation, we will refer to ($\mathcal{W}_{dae}$, $b_a$) as $\theta_{dae}$, ($\mathcal{W}_{daa_1}$, $\mathcal{W}_{daa_2}$, $b_c$, $b_d$) as $\theta_{daa}$, and ($\mathcal{W}_{disc_1}$, $\mathcal{W}_{disc_2}$, $b_e$, $b_f$) as $\theta_{disc}$.}.

Samples are allocated in a permuted paired sliding window to handle {\it contrasting throughput}. Training is performed on pairs of source and target samples by standard stochastic gradient descent of Equation \eqref{eq:loss-acdc}.

The DAE component uses a generative loss function in terms of the mean-squared error to handle the challenge of {\it different feature space} between the streams. For every pair of input samples, it performs a one-epoch greedy-layer wise pre-training (\cite{Bengio_Greedy}) without noise, followed by standard tied-weight training with a $10\%$ masking noise.

Meanwhile, both DAA and DISC modules use the multi-class logarithmic loss as its training function, with DAA performing a gradient reversal layer (GRL) (\cite{GRL}) transformation at $\theta_{dae}$.

\subsection{Module adaptation}\label{sec:module_adaptation}

\iffalse
\begin{figure}[ht]
\centering
\centerline{\includegraphics[width=\columnwidth]{images/node_evolution.png}}
\caption{Structural evolution on a USPS$\rightarrow$MNIST experiment.}
\label{fig:node-evolution}
\end{figure}
\fi

ACDC adopts a data-driven self-evolving structure driven by its active drift detector, adapting itself according to the data stream to handle {\it asynchronous drift}. It analyses the reconstruction error in the DAE module and the discriminative error on both DAA and DISC modules. It approximates the model's generalization power via its bias-variance decomposition, signalizing high bias (under-fitting) and high variance (over-fitting) situations.

ACDC uses the sigmoid function $\sigma ()$ for all its activation functions, making it easier to approximate the model's generalization by a probit function $\Phi (\xi \textbf{X}) = \int_{-inf}^{\textbf{X}} \mathcal{N}(\theta|0,1)d\theta$, with $\xi = \pi/8$ (\cite{ML:Prob}).

\begin{equation}\label{eq:expected_y_dae}
    E[\hat{y}]_{DAE} = \sigma \Bigg( \frac{\mu}{\sqrt{1 + \frac{\pi \sigma ^2}{8}}}\mathcal{W}_{dae} + b_{a} \Bigg)\mathcal{W}_{dae}^T + b_{b}
\end{equation}

\begin{equation}
\label{eq:expected_y_da}
    E[\hat{y}]_{DAA} = \sigma \Bigg(\sigma \Big( \frac{\mu}{\sqrt{1 + \frac{\pi \sigma ^2}{8}}}\mathcal{W}_{dae} + b_{a} \Big)\mathcal{W}_{daa_1} + b_{c}\Bigg)\mathcal{W}_{daa_2} + b_{d}
\end{equation}

\begin{equation}
\label{eq:expected_y_disc}
    E[\hat{y}]_{DISC} = \sigma \Bigg(\sigma \Big( \frac{\mu}{\sqrt{1 + \frac{\pi \sigma ^2}{8}}}\mathcal{W}_{dae} + b_{a} \Big)\mathcal{W}_{disc_1} + b_{e}\Bigg)\mathcal{W}_{disc_2} + b_{f}
\end{equation}

\noindent where $\mu$ and $\sigma^2$ are respectively the mean and variance of the samples fed, and $\hat{y}$ is the expected output. The integral of the probit function is then used to measure the module's bias and variance:

\begin{equation}\label{eq:ns-pre-1}
    Bias + Var = (E[\hat{y}]^2 + y) + (E[\hat{y}^2] + E[\hat{y}]^2)
\end{equation}

The conditions for growing and pruning follow a modified statistical process control (SPC) algorithm (\cite{gama2010knowledge, Gama04learningwith}):

\begin{equation}\label{eq:grow-prune-conditions-1}
    \beta = \alpha_1(-Bias)+\alpha_2
\end{equation}

\begin{equation}\label{eq:grow-prune-conditions-2}
    \Lambda = 2 * (\alpha_1(-Var)+\alpha_2)
\end{equation}

\begin{equation}\label{eq:grow-prune-conditions-3}
    \text{Growing condition:}\; \mu_{Bias} + \sigma_{Bias} \geq \mu_{Bias}^{min} + \beta * \sigma_{Bias}^{min}
\end{equation}

\begin{equation}\label{eq:grow-prune-conditions-4}
    \text{Pruning condition:}\; \mu_{Var} + \sigma_{Var} \geq \mu_{Var}^{min} + \Lambda * \sigma_{Var}^{min}
\end{equation}

The setting of $\beta$ and $\Lambda$ enables flexible growing and pruning conditions achieving a confidence interval in the range of $[\mu \pm \sigma, \mu \pm 2\sigma]$ and $[\mu \pm \sigma, \mu \pm 3\sigma]$, for the conditions for growing and pruning respectively.
This setting results in a node growing process sensitive to the high-bias case, while the node pruning process is reactive to the high-variance situation. It is worth noting $\Lambda$ is doubled in comparison to $\beta$ to avoid the  \textit{direct-pruning-after-growing} circumstance.
$\mu_{Bias}^{min}$, $\sigma_{Bias}^{min}$, $\mu_{Var}^{min}$, $\sigma_{Var}^{min}$ are re-initialized if the growing or pruning conditions are met.

Each module adapts independently, constantly growing or pruning the last layer before their output layer, as described by the color scheme in Figure \ref{fig:acdc-architecture}.

If the growing condition is met, a single new node initialized by Xavier's initialization (\cite{xavier}) is added to the respective module, increasing its capacity and reducing its bias. The network bias should decrease or at least become stable when there is no drift.

On the other hand, if the pruning condition is true, the weakest hidden node $r^*$ according to its expected activation degree is discarded:

\begin{equation}
    r^* = \underset{r=1,...,\mathcal{R}}{min} \sigma \Bigg( \frac{\mu^{r}}{\sqrt{1 + \pi \frac{\sigma^{2^{r}}}{8}}}\mathcal{W}_{in}^{r} + b_{in}^{r} \Bigg)
\end{equation}

\noindent where $\mathcal{R}$ is the number of nodes in the module's last hidden layer. The pruning operation aims to attenuate over-fitting by reducing the module capacity. Note that a small $r^*$ value indicates a hidden node that plays a small role in producing the module output and can be discarded without significant loss of accuracy.

As each ACDC module has a different objective function, each triggers the growing and pruning conditions on different moments, reacting to different levels of drifts. Furthermore, the DAE module is shared between the whole ACDC structure, so every change on its hidden layer causes direct interference to DAA and DISC. With this in mind, ACDC presents the optional hyper-parameter $\kappa$, which controls the number of internal epochs ACDC can carry out. ACDC still performs its evaluation under the prequential test-then-train protocol, never re-visiting a sample after its batch is over; however, during the same batch, the first $\kappa$ is considered a \textit{drift detection and adaptation} internal epoch while following $\kappa$ are considered pure \textit{learning} iterations, allowing the network to better learn under the new structure. A batch is a collection of $N_m$ samples that are accordingly allocated into sliding windows $W_{\mathcal{S}}$ and $W_{\mathcal{T}}$, i.e., $W_{\mathcal{S}}$ and $W_{\mathcal{T}}$ can have different sizes that sum to $N_m$.

Additionally, DAA's objective function allows it to detect asynchronous drift more subtly. ACDC uses this for its advantage, signalizing DISC to grow a node every time the DAA growing condition is activated. This procedure enables DISC to be more susceptible for source and target distributions changes, as we will see later in the ablation studies. If a forced growing from DAA into DISC causes any harm to the latter, DISC will soon overfit and auto-correct by its pruning condition activation.

Finally, regarding the initial number of hidden nodes, both DAA and DISC modules initialize their hidden layer with only one node, while DAE initializes its hidden layer with $u/2$ nodes, where $u$ is the feature vector size. Therefore, we arbitrarily choose $u/2$ as a general rule of thumb that provides enough compression capacity and learning speed to a denoising autoencoder.

\subsection{Cross-domain adaptation}

Many approaches bound the target error by the sum of the source error and a notion of distance between the source and the target distributions to tackle the challenging domain adaptation problem as demonstrated by Equation \eqref{eq:ben-david-1}. In this paper, we focus on the $\mathcal{H}$-divergence (\cite{ben2010theory}):

\textbf{Definition 1 (\cite{ben2010theory})}:  \textit{Given two domain distributions $\mathcal{D}_\mathcal{S}^\mathcal{X}$ and $\mathcal{D}_\mathcal{T}^\mathcal{X}$ over $\mathcal{X}$, and a hypothesis class $\mathcal{H}$, the $\mathcal{H}$-divergence between $\mathcal{D}_\mathcal{S}^\mathcal{X}$ and $\mathcal{D}_\mathcal{T}^\mathcal{X}$ is}\label{df-1}

\begin{equation}
    d_{\mathcal{H}}(\mathcal{D}_\mathcal{S}^\mathcal{X},\mathcal{D}_\mathcal{T}^\mathcal{X})\! = \! \underset{\eta \in \mathcal{H}}{2sup}\left|\underset{x \sim \mathcal{D}_\mathcal{S}^\mathcal{X}}{Pr[\eta(x) \! = \! 1]} \! - \! \underset{x \sim \mathcal{D}_\mathcal{T}^\mathcal{X}}{Pr[\eta(x) \! = \! 1]}  \right|
\end{equation}

\begin{equation}\label{eq:theorem-solution}
    \frac{\hat{d}_\mathcal{H}(\mathcal{S},\mathcal{T})}{2} = 1 - \underset{\eta \in \mathcal{H}}{min}\Big[\frac{1}{n}\sum_{i=1}^n I[\eta(x_i) = 0] + \frac{1}{N-n} \sum_{i=n+1}^{N} I[\eta(x_i) = 1]\Big]
\end{equation}

We assume in definition 1 that the hypothesis class $\mathcal{H}$ is a (discrete or continuous) set of binary classifiers $\eta : \mathcal{X} \rightarrow [0,1]$, which represents the source and target domain streams. The $\mathcal{H}$-divergence relies on the capacity of the hypothesis class $\mathcal{H}$ to distinguish between examples generated by $\mathcal{D}_\mathcal{S}^\mathcal{X}$ from examples generated by $\mathcal{D}_\mathcal{T}^\mathcal{X}$. For a symmetric hypothesis class $\mathcal{H}$, one can compute the  \textit{empirical $\mathcal{H}$-divergence} between two samples $\mathcal{S} = \{(x_i, y_i)_{i=1}^n \} \sim (\mathcal{D}_\mathcal{S}^{\mathcal{X}})^n$ and $\mathcal{T} = \{x_i\}_{i=n+1}^N \sim (\mathcal{D}_\mathcal{T}^\mathcal{X})^{N-n}$ by computing Equation \eqref{eq:theorem-solution} where $I[a]$ is 1 if $a$ is true, and $0$ otherwise (\cite{ben2010theory}).

Our approach handles {\it scarcity of labeled samples} in the target stream and {\it covariate shift} by creating a dynamic domain-invariance network. To learn a model that generalize well from one domain to another, we ensure that the internal representation of the neural network contains no discriminative information about the original domain of the input. We follow a simplified version of (\cite{DANN}) to estimate Equation \eqref{eq:theorem-solution} by a  \textrm{domain-adversarial classifier} that learns a logistics regressor $\text{DAA} : {\rm I\!R}^\mathcal{D} \rightarrow [0,1]$, which can be explained as in Equation \eqref{eq:full-cross-domain-loss}\footnote{Note that this explanation does not includes $\mathcal{L}_{DAE}$. The DAE module performs feature extraction on the streams, not being {\bf directly} linked to the cross-domain adaptation.} where we are seeking the parameters $\hat{\theta}_{dae}$, $\hat{\theta}_{disc}$, $\hat{\theta}_{daa}$.

\begin{equation}
    \label{eq:full-cross-domain-loss}
    E(\theta_{dae}, \theta_{disc}, \theta_{da}) = \frac{1}{n}\sum_{i=1}^n \mathcal{L}_{DISC}^i(\theta_{dae}, \theta_{disc}) - \frac{1}{n}\sum_{i=1}^n \hat{\mathcal{L}}_{DA}^i(\theta_{dae}, \theta_{da}) - \frac{1}{N-n} \sum_{i=n+1}^N \hat{\mathcal{L}}_{DA}^i(\theta_{dae}, \theta_{da})
\end{equation}

\begin{equation}
    \label{eq:domain_adaptation_optimization_min}
    (\hat{\theta}_{dae}, \hat{\theta}_{disc}) = \underset{\theta_{dae}, \theta_{disc}}{argmin}\: E(\theta_{dae}, \theta_{disc}, \hat{\theta}_{daa})
\end{equation}

\begin{equation}
    \label{eq:domain_adaptation_optimizaton_max}
    \hat{\theta}_{daa} = \underset{\theta_{daa}}{argmax}\: E(\hat{\theta}_{dae}, \hat{\theta}_{disc}, \theta_{daa})
\end{equation}

Thus, the optimization problem involves a minimization concerning some parameters, as well as maximization for others.

\section{Experiments}

ACDC's performance is evaluated using a combination of 11 datasets that create 26 distinct experiments. Finally, an ablation study further explores the impact of each of ACDC modules in the online cross-domain adaptation problem.

\subsection{Setup}

All the methods have been evaluated using a Windows 10 machine with a Intel Core i9-9900K 5.0 GHz with 32GB of main memory. The baselines are compared against 3 ACDC's instances, with $\kappa$ set to $\{1,3,5\}$, each containing the arbitrary values size of sliding window $N_m = 1000$, learning rate $\alpha = 0.01$ and momentum rate $\eta = 0.95$. Finally, $\alpha_1=1.25$ and $\alpha_2=0.75$ for all experiments, except CIFAR10$\leftrightarrow$STL10, where $\alpha_1=1.45$ and $\alpha_2=0.95$, allowing a suitable over-fitting/under-fitting control under a dense feature space.

Meanwhile, the baselines are executed following the default configurations reported in their own papers, with a minor modification on MSC and FUSION, where they receive only 10 samples for their warm-up stage. These settings were put in front after several tests to ensure that the baselines are not at a disadvantage compared to ACDC.

\subsection{Benchmarks}

\begin{table}[ht]
\begin{center}
\begin{tabular}{lrrr}
\toprule
Dataset               & Features & Classes & Samples     \\
\midrule
MNIST(MN)              & 784      & 10      & 70,000  \\
USPS(US)               & 256      & 10      & 9,298   \\
CIFAR10(CF)            & 512      & 9       & 54,000  \\
STL10(ST)              & 512      & 9       & 11,700  \\
London Bike(LD)        & 8        & 2       & 17,414  \\
Washington Bike(WA)    & 8        & 2       & 18,110      \\
Amazon@Beauty(AM1)     & 300      & 5       & 5,150   \\
Amazon@Books(AM2)      & 300      & 5       & 500,000 \\
Amazon@Industrial(AM3) & 300      & 5       & 73,146  \\
Amazon@Luxury(AM4)     & 300      & 5       & 33,784  \\
Amazon@Magazine(AM5)   & 300      & 5       & 2,230   \\
\bottomrule
\end{tabular}
\end{center}
\caption{Datasets' characteristics.}
\label{tb_dataset}
\end{table}

Table \ref{tb_dataset} lists the datasets used in the experiments. All datasets are publicly available real-world datasets.

\textbf{MNIST (MN) $\leftrightarrow$ USPS (US)}: Gray-scale images of hand-written digits collected from different sources sharing 10 classes. The USPS dataset (\cite{USPS}) consists of 9,298 images of size 16x16, while the MNIST dataset (\cite{lecun-mnisthandwrittendigit-2010}) consists of 70,000 images of size 28x28. We uniformly resize all the images to 16x16 in the US$\rightarrow$MN experiment and 28x28 in the MN$\rightarrow$US experiment.

\textbf{CIFAR10 (CF) $\leftrightarrow$ STL10 (ST)}: Full colored images used to train recognition models. One non-overlapping class is removed from both datasets. The CIFAR10 dataset (\cite{cifar10}) contains 54,000 workable images of size 32x32,  while the STL10 (\cite{stl10}) contains 11,700 workable images of size 96x96. All samples are fed into an ImageNet-pre-trained ResNet-18 (\cite{resnet}) to extract features from the colored images.

\textbf{Amazon@X (AM)}: A multi-domain sentiment dataset containing product reviews taken from Amazon.com where X denotes the product type (\cite{AmazonReview}). We arbitrarily choose 5 product types, obtaining two products with similar contexts and a non-related topic. To extract features from raw review text, we used the averaged summed output from Google's word2vec model pre-trained on 100 billion words (\cite{word2vec}).

\textbf{London (LD) $\leftrightarrow$ Washington (WA)}: Tabular data describing bike-sharing behaviors in the city of London (\cite{london-bike-sharing}) and Washington D.C. (\cite{washington-bike-sharing}). We pre-processed the features, so they represent the same information\footnote{Example: instead of having binary features "{\it weekday}" in one dataset and "{\it weekend}" in another dataset, a pre-process was put into place, so the features represent the same information in both datasets}.

Except for LD$\leftrightarrow$WA, which present a real drift due to its nature, we dynamically simulate abrupt concept drifts in these real-world datasets using a scaling hyper-plane strategy. After a time instance, every data point becomes
$x_i = (d_z \times x_i)/||x||$, where $d_z : {\rm I\!R}^u \rightarrow [0,...2)$ is a random generated concept drift vector. $z$ is the number of concepts drifts in the stream, where $z=5$ for every source stream and $z=7$ for every target stream, forcing the ascension of asynchronous drift during sample throughput\footnote{$z=5$ and $z=7$ are prime numbers, so artificial asynchronous drifts are guaranteed}. Finally, $d_1 : {\rm I\!R}^u \rightarrow 1$ for both source and target, while adjacent $d_z$ are generated with fixed pseudo-random seeds to guarantee a fair comparison between all baselines.

\subsection{Baselines}

We selected the set of baselines which made their source-code available online: ATL (\cite{ATL}) FUSION (\cite{FUSION}), Melanie (\cite{Melanie}), and MSC (\cite{MSC}). We re-evaluated their source codes under the same computation environment for a fair comparison, an Intel Core i9-9900K CPU with 32 GB of main memory. ACDC was built with Python 3, MSC and FUSION with Python 2, ATL with Matlab, and Melanie with Java, as an extension for the MOA framework (\cite{MOA-framework}). Their codes were modified to handle contrasting throughput.

To simulate a contrasting throughput data-stream setting into the benchmarks, ACDC and every baseline satiate with a ratio of source to target samples of 
$(n_\mathcal{S}-n_{W_\mathcal{S}})/(n_\mathcal{S}+n_\mathcal{T}-n_{W_\mathcal{S}}-n_{W_\mathcal{T}})$, where $n_{\mathcal{S}}$, $n_{\mathcal{T}}$, $n_{W_\mathcal{S}}$, $n_{W_\mathcal{T}}$ corresponds to the total amount of source and target samples, and the amount of source and target samples already received, respectively. The incoming ratio continually update after every incoming sample.

\subsection{Numerical results}

\begin{table*}[ht]
\begin{center}
\resizebox{\columnwidth}{!}{%
\begin{tabular}{cc|rrrr|rrr}
\toprule
\multicolumn{2}{c|}{Experiment} & \multirow{2}{*}{MSC (\%)} & \multirow{2}{*}{ATL (\%)} & \multirow{2}{*}{FUSION (\%)} & \multirow{2}{*}{Melanie (\%)} & \multirow{2}{*}{\textbf{ACDC-1} (\%)} & \multirow{2}{*}{\textbf{ACDC-3} (\%)} & \multirow{2}{*}{\textbf{ACDC-5} (\%)} \\
Source & Target & & & & & & \\

\midrule
\midrule

MN & US &  46.83 $\pm$ 0.29 & 54.18 $\pm$ 4.50 & 24.39 $\pm$ 1.87 & 10.67 $\pm$ 0.00 & 53.42 $\pm$ 0.92 & \underline{56.57 $\pm$ 2.27} & \textbf{59.12 $\pm$ 4.62}\\ % MNIST -> USPS
US & MN & 09.19 $\pm$ 0.48 & 21.82 $\pm$ 5.51 & 16.49 $\pm$ 1.04 & 13.20 $\pm$ 0.00 & \underline{41.16 $\pm$ 1.54} & \textbf{48.21 $\pm$ 1.28} & \textbf{48.71 $\pm$ 1.14} \\ % USPS -> MNIST

\hline
\hline

CF & ST & \textbf{50.50 $\pm$ 1.57} & 17.02 $\pm$ 2.35 & 19.90 $\pm$ 1.47 & 11.22 $\pm$ 0.00 & 42.34 $\pm$ 4.43 & \underline{43.38 $\pm$ 2.53} & 36.73 $\pm$ 2.72 \\ % CIFAR -> STL
ST & CF & 36.60 $\pm$ 0.31 & 19.39 $\pm$ 0.30 & 17.80 $\pm$ 0.82 & 11.20 $\pm$ 0.00 & \textbf{44.00 $\pm$ 1.66} & \underline{42.86 $\pm$ 1.26} & 37.79 $\pm$ 1.50 \\ % STL -> CIFAR

\hline
\hline

LD & WA & 65.73 $\pm$ 0.28 & 64.25 $\pm$ 0.96 & 63.05 $\pm$ 2.43 & 63.01 $\pm$ 0.02 & \underline{66.91 $\pm$ 0.01} & \textbf{69.73 $\pm$ 0.01} & \textbf{69.35 $\pm$ 0.01} \\ % London -> Washington
WA & LD & 60.17 $\pm$ 0.06 & 65.06 $\pm$ 0.20 & 51.92 $\pm$ 0.73 & 62.15 $\pm$ 0.01 & 64.49 $\pm$ 0.01 & \textbf{66.22 $\pm$ 0.01} & \underline{65.62 $\pm$ 0.02} \\ % Washington -> London

\hline
\hline

AM1 & AM2 & \textbf{62.56 $\pm$ 0.04} & \underline{59.37 $\pm$ 0.45} & \textbf{62.60 $\pm$ 0.01} & 14.71 $\pm$ 0.01 & \textbf{62.55 $\pm$ 0.25} & \textbf{62.41 $\pm$ 0.21} & \textbf{62.62 $\pm$ 0.12} \\ % Beauty -> Books
 & AM3 & \textbf{72.50 $\pm$ 0.03} & \underline{69.27 $\pm$ 1.84} & \textbf{72.53 $\pm$ 0.00} & 26.58 $\pm$ 0.01 & \textbf{72.98 $\pm$ 0.00} & \textbf{72.98 $\pm$ 0.00} & \textbf{72.98 $\pm$ 0.00} \\ % Beauty -> Industrial
 & AM4 & \underline{57.78 $\pm$ 0.01} & 55.88 $\pm$ 2.13 & \underline{57.07 $\pm$ 0.66} & 9.52 $\pm$ 0.02 & \textbf{60.25 $\pm$ 0.02} & \textbf{60.24 $\pm$ 0.02} & \textbf{60.25 $\pm$ 0.02} \\ % Beauty -> Luxury
 & AM5 & 63.87 $\pm$ 0.59 & \underline{64.87 $\pm$ 0.26} & \underline{64.64 $\pm$ 0.00} & 5.60 $\pm$ 0.11 & \textbf{71.28 $\pm$ 0.15} & \textbf{71.11 $\pm$ 0.28} & \textbf{70.96 $\pm$ 0.23} \\ % Beauty -> Magazine

\hline

AM2 & AM1 & \underline{86.84 $\pm$ 1.24} & 68.98 $\pm$ 15.2 & \underline{86.06 $\pm$ 1.52} & 4.35 $\pm$ 0.00 & \textbf{88.34 $\pm$ 0.03} & \textbf{88.36 $\pm$ 0.02} & \textbf{88.33 $\pm$ 0.04} \\ % Books -> Beauty
 & AM3 & 67.29 $\pm$ 2.99 & 54.06 $\pm$ 9.93 & \underline{71.28 $\pm$ 0.14} & 6.89 $\pm$ 0.01 & \textbf{72.58 $\pm$ 0.00} & \textbf{72.58 $\pm$ 0.00} & \textbf{72.58 $\pm$ 0.00} \\ % Books -> Industrial
 & AM4 & 53.32 $\pm$ 0.61 & 47.50 $\pm$ 7.77 & \underline{54.11 $\pm$ 1.23} & 4.10 $\pm$ 0.00 & \textbf{57.95 $\pm$ 0.02} & \textbf{57.96 $\pm$ 0.01} & \textbf{57.95 $\pm$ 0.01} \\ % Books -> Luxury
 & AM5 & \underline{60.51 $\pm$ 2.99} & 58.48 $\pm$ 3.24 & \textbf{64.63 $\pm$ 0.00} & 3.62 $\pm$ 0.01 & \textbf{64.57 $\pm$ 0.07} & \textbf{64.60 $\pm$ 0.03} & \textbf{64.51 $\pm$ 0.03} \\ % Books -> Magazine
 
 \hline

AM3 & AM1 & \textbf{88.35 $\pm$ 0.00} & 86.92 $\pm$ 2.72 & \underline{87.78 $\pm$ 0.00} & 8.40 $\pm$ 0.00 & \textbf{88.78 $\pm$ 0.00} & \textbf{88.77 $\pm$ 0.00} & \textbf{88.77 $\pm$ 0.00} \\ % Industrial -> Beauty
 & AM2 & \underline{61.99 $\pm$ 0.1}5 & 52.83 $\pm$ 3.83 & \textbf{62.57 $\pm$ 0.02} & 13.39 $\pm$ 0.02 & \textbf{62.67 $\pm$ 0.00} & \textbf{62.67 $\pm$ 0.00} & \textbf{62.67 $\pm$ 0.00} \\ % Industrial -> Books
 & AM4 & \underline{57.81 $\pm$ 0.00} & 55.74 $\pm$ 1.91 & \underline{57.71 $\pm$ 0.04} & 5.67 $\pm$ 0.01 & \textbf{58.65 $\pm$ 0.02} & \textbf{58.63 $\pm$ 0.01} & \textbf{58.63 $\pm$ 0.03} \\ % Industrial -> Luxury 
 & AM5 & \underline{64.64 $\pm$ 0.00} & \underline{64.11 $\pm$ 0.27} & \underline{64.64 $\pm$ 0.00} & 3.65 $\pm$ 0.01 & \textbf{65.25 $\pm$ 0.12} & \textbf{65.15 $\pm$ 0.09} & \textbf{65.11 $\pm$ 0.03} \\ % Industrial -> Magazine
 
 \hline

AM4 & AM1 & 86.55 $\pm$ 1.41 & 73.65 $\pm$ 15.7 & \underline{88.35 $\pm$ 0.00} & 13.40 $\pm$ 0.01 & \textbf{89.08 $\pm$ 0.02} & \underline{88.76 $\pm$ 0.59} & \textbf{89.06 $\pm$ 0.02} \\ % Luxury -> Beauty
 & AM2 & 47.41 $\pm$ 1.72 & \underline{48.51 $\pm$ 2.10} & \textbf{62.49 $\pm$ 0.07} & 14.10 $\pm$ 0.00 & \textbf{62.28 $\pm$ 0.89} & \textbf{62.73 $\pm$ 0.00} & \textbf{62.73 $\pm$ 0.00} \\ % Luxury -> Books
 & AM3 & 56.22 $\pm$ 2.63 & \underline{59.94 $\pm$ 10.1} & \textbf{72.53 $\pm$ 0.00} & 20.46 $\pm$ 0.01 & \textbf{72.94 $\pm$ 0.21} & \textbf{73.05  $\pm$ 0.00} & \textbf{73.04 $\pm$ 0.00} \\ % Luxury -> Industrial
 & AM5 & \underline{64.44 $\pm$ 0.21} & 54.98 $\pm$ 5.25 & \underline{64.64 $\pm$ 0.00} & 3.90 $\pm$ 0.01 & \textbf{65.34 $\pm$ 0.08} & \textbf{65.40 $\pm$ 0.06} & \textbf{65.40 $\pm$ 0.09} \\ % Luxury -> Magazine
 
 \hline

AM5 & AM1 & \underline{84.87 $\pm$ 2.37} & 79.06 $\pm$ 1.35 & 22.32 $\pm$ 10.6 & 57.05 $\pm$ 0.05 & \textbf{90.72 $\pm$ 0.03} & \textbf{90.68 $\pm$ 0.04} & \textbf{90.66 $\pm$ 0.05} \\ % Magazine -> Beauty
 & AM2 & 43.12 $\pm$ 3.67 & 58.71 $\pm$ 2.62 & \underline{61.94 $\pm$ 0.14} & 14.80 $\pm$ 0.00 & 54.89 $\pm$ 14.9 & \underline{61.51 $\pm$ 1.39} & \textbf{62.33 $\pm$ 0.35} \\ % Magazine -> Books
 & AM3 & 52.51 $\pm$ 7.96 & \underline{71.65 $\pm$ 1.24} & 70.42 $\pm$ 1.35 & 27.36 $\pm$ 0.01 & \textbf{73.03 $\pm$ 0.00} & \textbf{73.04 $\pm$ 0.00} & \textbf{73.04 $\pm$ 0.00} \\ % Magazine -> Industrial
 & AM4 & 50.66 $\pm$ 4.72 & \underline{55.79 $\pm$ 2.23} & 51.66 $\pm$ 0.83 & 10.16 $\pm$ 0.01 & \textbf{58.89 $\pm$ 0.01} & \textbf{58.90 $\pm$ 0.00} & \textbf{58.90 $\pm$ 0.00} \\ % Magazine -> Luxury
\bottomrule
\end{tabular}%
}
\end{center}
\caption{Target accuracy comparison for 5 executions of each experiment.}
\label{tb_results}
\end{table*}

\begin{figure}[ht]
    \centering
    \includegraphics{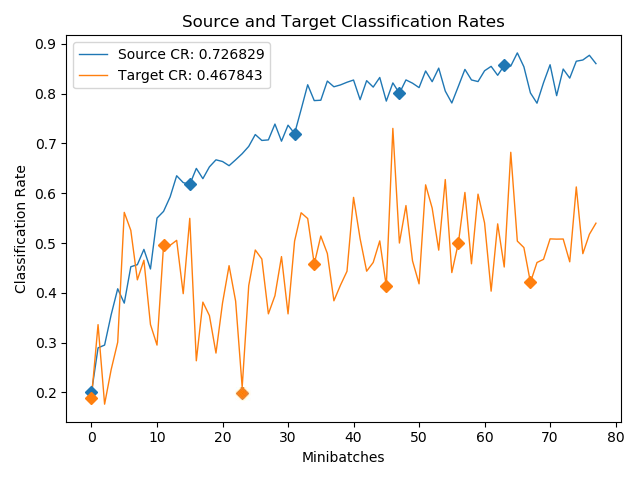}
    \caption{Source and Target accuracy evolution on experiment US$\rightarrow$MN and approximate locations of concept drifts.}
    \label{fig:acdc-time-classification-rate}
\end{figure}

Positive conclusions can be drawn from Table \ref{tb_results}, to which ACDC is capable of achieving the highest results of target classification rate in every experiment, besides CF$\rightarrow$ST, most of the time outperforming the baselines. Besides Melanie, the baselines present a good performance on several benchmarks, but they cannot consistently deliver excellent scores. Furthermore, MSC performance on CF$\rightarrow$ST is a surprise, especially when FUSION, a model with several similarities to the former, performs poorly on the same benchmark.

To add to these observations, Figure \ref{fig:acdc-time-classification-rate} presents the evolution of ACDC's source and target classification rates for the MN$\rightarrow$US experiment. Although we cannot pinpoint precisely when concept drift occurs due to the random nature of incoming samples, we know that the source stream was split into 5 concepts while the target stream was divided into 7 streams by our controlled experiment. Thus, considering the source stream classification rate, we can easily perceive that ACDC quickly recovers from concept drifts. Furthermore, this analysis is suitable because the source stream is labeled, so we mainly analyze DISC's adaptation performance with little interference from DAA.

CF$\leftrightarrow$ST experiments are interesting cases. Their feature spaces are too dense after ResNet-18 being applied, so we had to increase $\alpha_1$ and $\alpha_2$, parameters of the modified SPC algorithm. However, although ACDC can usually grow and prune nodes when training it, it over-fits with an increase of $\kappa$, showing a deteriorating performance under the prequential test-then-train protocol. 

ACDC's iteration across internal epochs occurs per data chunk of size $N_m$ rather than the whole data set, satisfying the online learning requirements, as processed batches are discarded and not revisited. An increase in the number of internal epochs $\kappa$, however, does not guarantee an increase in classification rate performance, as seen in the majority of the {\it Amazon@X} experiments, and lightly on WA$\leftrightarrow$LD, problems that come with a significant amount of natural noise.

Additionally, MSDA and COMC (\cite{onlineHeterogeneousTexas, COMC}) were not included into Table \ref{tb_results} because their source-codes are not publicly available. However, we can take in consideration the overlapping experiments US$\rightarrow$MN results published on their paper: COMC achieves a target accuracy(\%) of $42.37$, while MSDA achieves $28.72 \pm 0.56$ and $29.96 \pm 0.64$. Concluding, ACDC still outperform them.

\subsection{Ablation study}

We performed ablation study under four different configurations on the MN$\leftrightarrow$US experiments and $\kappa = 1$:
\renewcommand{\theenumi}{\Alph{enumi}}
\begin{enumerate}
    \item DAA is deactivated; 
    \item No self-evolving mechanism. Hidden-layers contain 100 hidden nodes;
    \item DAE starts with a single node;
    \item DAA's ability to grow nodes in DISC is turned off.
\end{enumerate}

\begin{table}[ht!]
    \centering
    \begin{tabular}{c|cc}
    \toprule
         Ablation & MN$\rightarrow$US & US$\rightarrow$MN \\
         \midrule
         \midrule
         A(\%) & 38.74 $\pm$ 4.12 & 27.86 $\pm$ 3.68 \\
         B(\%) & 56.22 $\pm$ 1.88 & 37.10 $\pm$ 3.85 \\
         C(\%) & 20.22 $\pm$ 1.44 & 20.38 $\pm$ 0.53 \\
         D(\%) & 48.84 $\pm$ 3.00 & 38.82 $\pm$ 1.43 \\
         \bottomrule
    \end{tabular}
    \caption{Performance for 5 executions of ablation study.}
    \label{tb_ablation}
\end{table}

\begin{table*}[ht!]
\begin{center}
\resizebox{\columnwidth}{!}{%
\begin{tabular}{cc|rrrr|rrr}
\toprule
\multicolumn{2}{c|}{Experiment} & \multirow{2}{*}{MSC (s)} & \multirow{2}{*}{ATL (s)} & \multirow{2}{*}{FUSION (s)} & \multirow{2}{*}{Melanie (s)} & \multirow{2}{*}{\textbf{ACDC-1} (s)} & \multirow{2}{*}{\textbf{ACDC-3} (s)} & \multirow{2}{*}{\textbf{ACDC-5} (s)} \\
Source & Target & & & & & \\

\midrule
\midrule

MN & US &  8,498 $\pm$ 395   & 66,160 $\pm$ 3,210 & 110 $\pm$ 3 & 1,002 $\pm$ 41 & 4,635 $\pm$ 12  & 13,199 $\pm$ 1,160 & 23,258 $\pm$ 1,302 \\ % MNIST -> USPS
US & MN & 11,587 $\pm$ 9,111 & 346 $\pm$ 5        & 115 $\pm$ 4 & 555 $\pm$ 26   & 1,917 $\pm$ 109 & 3,299  $\pm$ 25    & 4,926 $\pm$ 165 \\ % USPS -> MNIST

\hline
\hline

CF & ST & 5,900 $\pm$ 181  & 1,721 $\pm$ 245  & 95 $\pm$ 6  & 559 $\pm$ 64    & 1,536 $\pm$ 169 & 3,636 $\pm$ 290 & 4,945 $\pm$ 109 \\ % CIFAR -> STL
ST & CF & 11,437 $\pm$ 245 & 18,323 $\pm$ 171 & 147 $\pm$ 9 & 1,181 $\pm$ 119 & 1,564 $\pm$ 119 & 3,381 $\pm$ 280 & 4,580 $\pm$ 280 \\ % STL -> CIFAR

\hline
\hline

LD & WA & 2,789 $\pm$ 40  & 254 $\pm$ 3 & 30 $\pm$ 0 & 2 $\pm$ 0  & 172 $\pm$ 7  & 176 $\pm$ 1  & 175 $\pm$ 2 \\ % London -> Washington
WA & LD & 3,124 $\pm$ 109 & 262 $\pm$ 3 & 31 $\pm$ 0 & 2 $\pm$ 0  & 169 $\pm$ 13 & 167 $\pm$ 10 & 152 $\pm$ 10 \\ % Washington -> London

\hline
\hline

AM1 & AM2 & 58,683 $\pm$ 1,846 & 16,936 $\pm$ 601 & 960 $\pm$ 29 & 1,243 $\pm$ 100 & 12,869 $\pm$ 561 & 20,780 $\pm$ 1,447 & 27,918 $\pm$ 2,880 \\ % Beauty -> Books
    & AM3 & 13,282 $\pm$ 1,678 & 238 $\pm$ 6      & 150 $\pm$ 18 & 144 $\pm$ 4     & 1,923 $\pm$ 78   & 3,824 $\pm$ 99     & 5,649 $\pm$ 53 \\ % Beauty -> Industrial
    & AM4 & 6,730 $\pm$ 1,480  & 201 $\pm$ 2      & 68 $\pm$ 2   & 59 $\pm$ 2      & 870 $\pm$ 37     & 1,685 $\pm$ 74     & 2,497 $\pm$ 185 \\ % Beauty -> Luxury
    & AM5 & 787 $\pm$ 43       & 13 $\pm$ 1       & 11 $\pm$ 0   & 8 $\pm$ 0       & 120 $\pm$ 2      & 242 $\pm$ 2        & 361 $\pm$ 3 \\ % Beauty -> Magazine
 
 \hline

AM2 & AM1 & 39,527 $\pm$ 3,411 & 11,683 $\pm$ 527   & 531 $\pm$ 37 & 529 $\pm$ 13 & 11,305 $\pm$ 158   & 22,932 $\pm$ 1,957 & 41,645 $\pm$ 10,019 \\ % Books -> Beauty
    & AM3 & 58,406 $\pm$ 4,711 & 17,996 $\pm$ 1,223 & 729 $\pm$ 55 & 763 $\pm$ 14 & 11,424 $\pm$ 293   & 21,529 $\pm$ 606   & 38,908 $\pm$ 6,019 \\ % Books -> Industrial
    & AM4 & 45,749 $\pm$ 3,392 & 13,705 $\pm$ 792   & 586 $\pm$ 36 & 646 $\pm$ 10 & 15,512 $\pm$ 2,228 & 22,563 $\pm$ 1,296 & 35,274 $\pm$ 4,743 \\ % Books -> Luxury
    & AM5 & 34,540 $\pm$ 818   & 11,393 $\pm$ 403   & 552 $\pm$ 34 & 554 $\pm$ 7  & 17,448 $\pm$ 5,242 & 23,634 $\pm$ 2,335 & 37,271 $\pm$ 3,544 \\ % Books -> Magazine
 
 \hline

AM3 & AM1 & 7,220 $\pm$ 551    & 85 $\pm$ 2       & 90 $\pm$ 8     & 137 $\pm$ 1    & 1,965 $\pm$ 89   & 3,853 $\pm$ 165    & 5,909 $\pm$ 43 \\ % Industrial -> Beauty
    & AM2 & 58,631 $\pm$ 1,127 & 16,961 $\pm$ 282 & 1,067 $\pm$ 66 & 1,602 $\pm$ 47 & 11,545 $\pm$ 358 & 27,726 $\pm$ 2,100 & 43,329 $\pm$ 6,129 \\ % Industrial -> Books
    & AM4 & 12,294 $\pm$ 630   & 170 $\pm$ 3      & 143 $\pm$ 12   & 179 $\pm$ 12   & 1,639 $\pm$ 19   & 3,081 $\pm$ 30     & 4,571 $\pm$ 36 \\ % Industrial -> Luxury 
    & AM5 & 6,708 $\pm$ 398    & 74 $\pm$ 2       & 82 $\pm$ 6     & 116 $\pm$ 2    & 1,644 $\pm$ 13   & 3,300 $\pm$ 256    & 5,524 $\pm$ 320 \\ % Industrial -> Magazine
 
 \hline

AM4 & AM1 & 3,281 $\pm$ 163    & 119 $\pm$ 2      & 47 $\pm$ 2   & 53 $\pm$ 0     & 1,090 $\pm$ 79   & 2,077 $\pm$ 264    & 3,077 $\pm$ 151 \\ % Luxury -> Beauty
    & AM2 & 53,670 $\pm$ 2,892 & 16,678 $\pm$ 607 & 974 $\pm$ 58 & 1,398 $\pm$ 72 & 15,087 $\pm$ 722 & 27,914 $\pm$ 1,308 & 37,773 $\pm$ 5,947 \\ % Luxury -> Books
    & AM3 & 14,285 $\pm$ 1,132 & 350 $\pm$ 16     & 177 $\pm$ 11 & 172 $\pm$ 4    & 1,633 $\pm$ 20   & 3,226  $\pm$ 146   & 4,451 $\pm$ 326 \\ % Luxury -> Industrial
    & AM5 & 2,906 $\pm$ 192    & 132 $\pm$ 6      & 40 $\pm$ 2   & 46 $\pm$ 0     & 760 $\pm$ 15     & 1,449 $\pm$ 23     & 2,337 $\pm$ 172 \\ % Luxury -> Magazine
 
 \hline

AM5 & AM1 & 579 $\pm$ 22       & 17 $\pm$ 0       & 13 $\pm$ 0   & 12 $\pm$ 0      & 163 $\pm$ 36        & 369 $\pm$ 57       & 420 $\pm$ 69 \\ % Magazine -> Beauty
    & AM2 & 38,079 $\pm$ 2,262 & 14,433 $\pm$ 546 & 961 $\pm$ 43 & 1,376 $\pm$ 120 & 12,314 $\pm$ 1,001  & 21,116 $\pm$ 3,685 & 38,171 $\pm$ 1,390 \\ % Magazine -> Books
    & AM3 & 7,268 $\pm$ 667    & 218 $\pm$ 3      & 137 $\pm$ 5  & 150 $\pm$ 4     & 1,807 $\pm$ 99      & 3,637 $\pm$ 269    & 5,459 $\pm$ 438 \\ % Magazine -> Industrial
    & AM4 & 3,277 $\pm$ 259    & 172 $\pm$ 2      & 62 $\pm$ 2   & 54 $\pm$ 0      & 894 $\pm$ 53        & 1,587 $\pm$ 111    & 2,258 $\pm$ 236 \\ % Magazine -> Luxury
\bottomrule
\end{tabular}%
}
\end{center}
\caption{Training time comparison for 5 executions of each experiment.}
\label{tb_time_results}
\end{table*}

Table \ref{tb_ablation} depicts ACDC's ablation study numerical results, which shows the impact and importance of its primary design decisions. For example, in study (A), the absence of DAA drastically decreases the classification accuracy in both target and source domains by at least 5\%. 

Referring to study (B), a fixed structure adds extra hyper-parameter tunability, decreasing ACDC's plug-and-play factor. The adaptation process allows ACDC to slowly search for the optimal structure while training. In the MN$\rightarrow$US experiment, ACDC had a better performance than with an adaptive network because the model was already initialized with enough hidden nodes near the optimal structure size. In contrast, the US$\rightarrow$MN experiment had a decreasing result, as now ACDC has a fixed structure bigger and farther than the optimal structure.

The ablation study (C) highlights the importance of setting an initial capacity for DAE to handle data compression without information loss. DAE needs to have access to the right amount of information since its initialization, consequently allowing other modules to learn, as depicted in Figure \ref{fig:acdc-architecture}.

Lastly, ablation study (D) shows hows the interaction between DAA and DISC is fundamental for fast pacing learning.

\subsection{Space and Time complexity}

\begin{figure}[ht]
    \centering
    \begin{subfigure}{0.49\textwidth}
        \centering
        \includegraphics[width=\columnwidth]{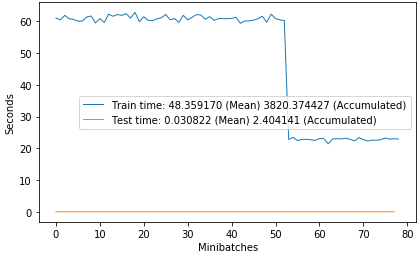}
        \caption{Processing time}
        \label{fig:acdc-time-space-complexity-1}
    \end{subfigure}
    \begin{subfigure}{0.49\textwidth}
        \centering
        \includegraphics[width=\columnwidth]{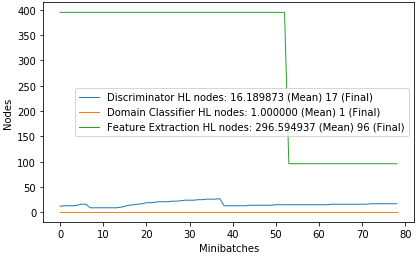}
        \caption{Hidden node evolution}
        \label{fig:acdc-time-space-complexity-2}
    \end{subfigure}
    \caption{Plots over sliding window for ablation study (A) on experiment MN$\rightarrow$US.}
\end{figure}

Table \ref{tb_time_results} gives an overall idea of ACDC and its baselines' time complexity in terms of total processing time. Time complexity is probably ACDC's weakest point. The adaptation procedure is an expensive operation performed sample by sample causing a big impact on its time complexity, even though ACDC operates on a sliding window. It is important to note that ACDC was operated on the CPU to have a fair comparison with its baselines; however, it is important to highlight that its current adaptive nature is not beneficial for GPUs either. Many operations with single samples are performed, which is disadvantageous for the paralleled architecture of a graphic processing unit.

Furthermore, it is valid to explore how ACDC's adaptation impacts its time and space complexity throughout the training process, not only as a final definitive metric. We choose to use one of the ablation studies to illustrate this phenomenon in Figures \ref{fig:acdc-time-space-complexity-1} and \ref{fig:acdc-time-space-complexity-2}, as it is easier to see the effect of the number of nodes in the network training time.

ACDC's modules grow nodes when underfitting and prune nodes when overfitting. Therefore, whenever ACDC creates a new node, it requires more memory and more computation power, increasing its space and time complexity. Similarly, whenever ACDC prune a node or a set of nodes, it will free some memory and computation requirement, decreasing its space and time complexity. ACDC's adaptation conditions are evaluated on every sample, which causes the fluctuation in training in time in Figure \ref{fig:acdc-time-space-complexity-1}. However, around chunk 52, we can see in Figure \ref{fig:acdc-time-space-complexity-2} that ACDC prunes several nodes from the DAE module, drastically decreasing the training time.

\subsection{Impact of internal epochs $\kappa$}

\begin{figure}[ht]
    \centering
    \begin{subfigure}{0.49\textwidth}
        \centering
        \includegraphics[width=\columnwidth]{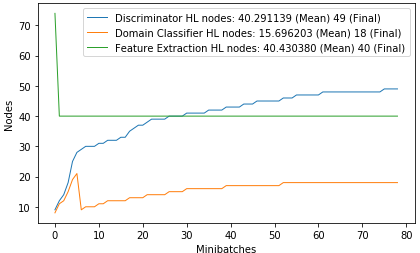}
        \caption{Internal epoch $\kappa = 1$}
        \label{fig:internal-epochs-1}
    \end{subfigure}
    \begin{subfigure}{0.49\textwidth}
        \centering
        \includegraphics[width=\columnwidth]{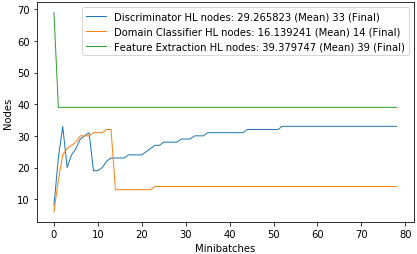}
        \caption{Internal epoch $\kappa = 5$}
        \label{fig:internal-epochs-5}
    \end{subfigure}
    \caption{Node evolution plots over sliding window with different $\kappa$ on experiment US$\rightarrow$MN.}
\end{figure}

When taking the target accuracy in respect, some experiments benefit from a more significant training cycle, as depicted in Table \ref{tb_results}. Here we will explore how the ACDC behaves with more internal epochs $\kappa$ for parameter learning.

Figures \ref{fig:internal-epochs-1} and \ref{fig:internal-epochs-5} present two plots for the structural evolution over the US$\rightarrow$MN experiment, wherein Figure \ref{fig:internal-epochs-1} we have $\kappa = 1$ and in Figure \ref{fig:internal-epochs-5} we have $\kappa = 5$. As a reminder, the module adaptation only occurs on the first internal epoch $\kappa$; however, the parameter learning stage happens in all internal epochs. With $\kappa > 1$, ACDC attempts to better learn that batch before receiving and processing the next batch. Therefore, it has opportunities to improve its parameter learning, with the risk of increasing its time complexity. When comparing Figures \ref{fig:internal-epochs-1} and \ref{fig:internal-epochs-5}, we can infer that experiments with $\kappa > 1$ has more relevant hidden-layers, as it is able to produce similar or better results for the US$\rightarrow$MN experiment with less hidden-nodes per module.

\section{Conclusion}

This paper proposes the autonomous cross-domain conversion (ACDC), an adversarial unsupervised cross-domain adaptation framework that uses a dynamic structure to react to data drifts actively. ACDC highlights three modules that encapsulate a solution to the online unsupervised cross-domain adaptation problem: a denoising autoencoder acting as a generative feature extractor, a domain-adversarial adaptation network performing cross-domain adaptation, and a discriminator.

We compared ACDC with a set of solid baselines under the prequential test-then-train protocol, yielding positive generalization and target accuracy conclusions. Additionally, ACDC achieved the highest result in almost every experiment, with an improvement of more than 10\% in some exceptional cases. We also explore the impact of ACDC dynamic structure on it space and time complexity, and how the internal epochs $\kappa$ influences the overall classifier generalization, usually providing more opportunities for the model to learn its parameter but with a high time complexity penalty cost.

ACDC is a flexible online neural network framework, hence having the opportunity to be future expanded to work well in the presence of convolution layers, distribute computing, recurrent layers, and variate layer depth, to say a few. Extensive experiments under the prequential test-then-train protocol on real-world data confirm that ACDC has significantly better performance in terms of target error rate than the baselines.

\bibliographystyle{unsrtnat}
\bibliography{main}
\end{document}